\documentclass[a4paper,10pt,twocolumn]{article}
\usepackage[utf8]{inputenc}
\usepackage{multicol,graphicx}
\usepackage{mathptmx}
\usepackage[T1]{fontenc}
\usepackage{textcomp}
\usepackage{url}
\usepackage{amssymb}

\usepackage[T1]{fontenc}
\usepackage[english]{babel}

\usepackage[backend=bibtex,style=ieee,doi=false,isbn=false,url=true,maxnames=6,citestyle=numeric-comp,giveninits=true]{biblatex}

\fontfamily{ptm}\selectfont

\bibliography{references.bib}

\usepackage[font=small]{caption}
\usepackage{amsmath}

\newcommand{\mysection}[1]{\vspace{0.4cm} \uppercase{#1} \vspace{0.4cm}}
\newcommand{\mysubsection}[1]{\hspace{10pt}\textit{#1:}}

\begin{document}
	
\setlength{\textfloatsep}{10pt plus 1.0pt minus 2.0pt}	
\setlength{\columnsep}{1cm}


\twocolumn[%
\begin{@twocolumnfalse}
\begin{center}
	{\fontsize{14}{18}\selectfont
        \textbf{{BENCHMARKING POSITIONAL ENCODING STRATEGIES FOR TRANSFORMER-BASED EEG FOUNDATION MODELS}}\\}
    \begin{large}
        \vspace{0.6cm}
        Ayşe Betül Yüce\textsuperscript{1}, Sebastian Stober\textsuperscript{1}\\
        \vspace{0.6cm}
        \textsuperscript{1}Department of Computer Science, Otto von Guericke University, Magdeburg, Germany\\
        \vspace{0.5cm}
        E-mail: ayse.yuece@ovgu.de
        \vspace{0.4cm}
    \end{large}
\end{center}	
\end{@twocolumnfalse}%
]%


ABSTRACT: Electroencephalography (EEG) is a widely used non-invasive technique for measuring brain activity in brain–computer interface (BCI) applications. Supervised EEG decoding models often struggle to generalize across tasks, subjects, and datasets, motivating transformer-based EEG foundation models trained with self-supervised learning. Since transformers are permutation-invariant, they require explicit positional information. Unlike textual tokens, EEG electrodes are spatially distributed across the scalp, raising the question of how electrode positions should be encoded in transformer-based EEG models.
In this study, we benchmark five positional encoding strategies within the CBraMod backbone and evaluate them under linear probing and fine-tuning protocols on motor imagery classification and emotion recognition. Our results show that no single strategy consistently outperforms across tasks. Spherical Positional Encoding (SPE) yields strong representations for motor imagery but underperforms on emotion recognition, while Asymmetric Conditional Positional Encoding (ACPE) demonstrates more consistent performance across tasks. These findings suggest that the optimal positional encoding strategy is task-dependent, with no universal solution across EEG decoding scenarios.


\mysection{introduction}



Electroencephalography (EEG) is a brain activity recording technique that provides high temporal resolution. Due to its non-invasive nature, it is widely preferred in both clinical\cite{fernandez2025eeg, gilavila2023discovereeg} and brain–computer interface (BCI) applications\cite{Guerrero-Mendez_2023, torres2020eegbciemotion}. However, EEG signals present substantial challenges. They exhibit a low signal-to-noise ratio and are highly susceptible to artifacts, session variability, and inter-subject differences. Furthermore, their high temporal resolution results in complex, high-dimensional data that require careful modeling. Early BCI systems relied on supervised learning with handcrafted features, which often generalize poorly across tasks, subjects, and datasets. These limitations motivate the development of scalable representation learning frameworks capable of learning transferable neural representations.

Inspired by advances in large-scale modeling in natural language processing\cite{bert2019} and computer vision\cite{vit}, recent studies have introduced transformer-based foundation models for EEG \cite{bendr2021, biot2023, cbramod2025, labram2024, neurogpt2024}. These models leverage self-supervised pretraining to learn general neural representations that can be adapted to downstream tasks. Since transformers operate through self-attention and are inherently permutation-invariant, they require explicit positional information to encode structural dependencies. Consequently, positional encoding (PE) becomes a critical architectural component. EEG electrodes follow anatomically meaningful spatial configurations that vary across datasets due to differences in channel number and electrode placement.



\begin{figure*}[t]
	\centering
	\includegraphics[width=\textwidth]{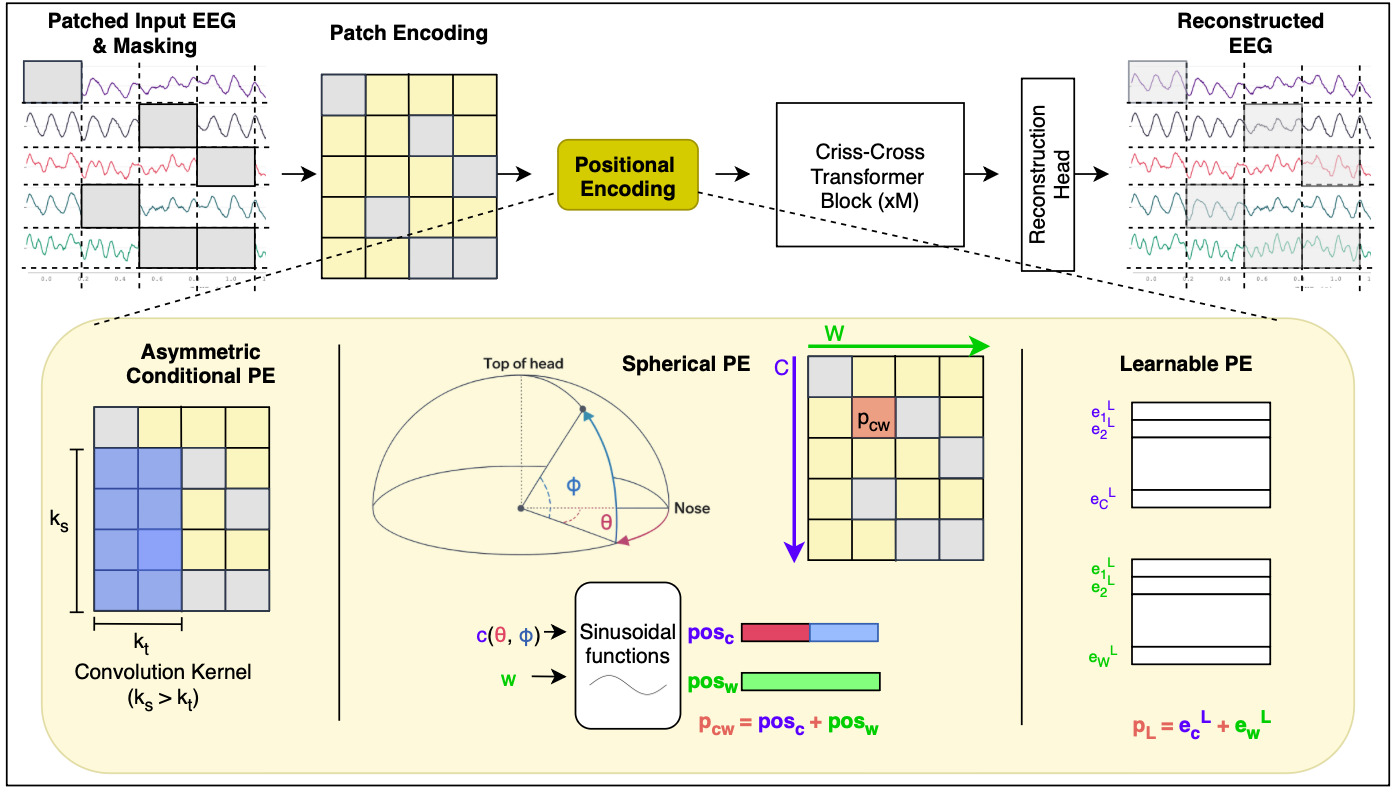}
	\caption{Overview of the experimental setup. The CBraMod backbone is retained with its original transformer blocks, while the positional encoding module is replaced with three primary variants under investigation: asymmetric conditional PE, Spherical PE, and learnable PE. Additionally, we extend SPE with a projection module and include a no positional encoding baseline.}
	\label{fig:block}
\end{figure*}

Recent transformer-based EEG foundation models adopt diverse mechanisms to incorporate positional information while accounting for the spatial–temporal structure of EEG signals. BIOT \cite{biot2023} addressed cross-dataset variability using channel-specific embeddings combined with relative temporal positional encoding. Subsequently, Wang J et al. \cite{cbramod2025} proposed an asymmetric conditional positional encoding (ACPE) module based on depthwise two-dimensional convolutions, where positional representations are dynamically generated from the spatio-temporal neighborhoods of EEG patches. By employing asymmetric kernels with a larger receptive field along the spatial dimension and a smaller one along the temporal dimension, their design captures long-range inter-channel dependencies while preserving short-range temporal context, enabling adaptation to varying EEG formats. Moving toward geometry-aware strategies, Ouahidi YE et al. \cite{reve2025} leveraged Cartesian $(x, y, z)$ electrode coordinates, augmented with noise to improve robustness to head-size and placement variations, and extended them with temporal indices to form a four-dimensional $(x, y, z, t)$ representation. The resulting 4D representation was projected into a multi-frequency Fourier space and combined with learned embeddings. Some researchers learned spatial and temporal embeddings during training and used them as positional encoding \cite{labram2024, klein}. These diverse implementations highlight the lack of systematic evaluation of positional encoding strategies in EEG foundation models. 


As part of this benchmark, we introduce Spherical Positional Encoding (SPE), which leverages the spherical geometry of electrode placement. For each electrode, azimuth and inclination angles are computed and transformed into fixed-length representations using sinusoidal functions at multiple frequencies, and the final embedding is obtained by concatenating these angular encodings. By encoding angular relationships rather than absolute Cartesian positions, SPE preserves intrinsic scalp topology while remaining invariant to global scaling and head-size differences, without introducing additional learnable parameters.

In this work, we systematically evaluate positional encoding strategies within the CBraMod foundation model, which employs a criss-cross transformer mechanism to capture temporal and spatial dependencies between EEG patches. As illustrated in Fig. \ref{fig:block}, we compare three primary strategies: ACPE, SPE, and learnable spatio-temporal PE. Additionally, we extend SPE with a projection module and include a no positional encoding baseline to further isolate the contribution of positional information. By keeping the backbone architecture fixed and modifying only the positional encoding component, we isolate the impact of positional design on representation learning and downstream performance. The contributions of this work are summarized as follows:

\begin{itemize}
    \item We conduct \textbf{a controlled benchmark of positional encoding strategies} in a transformer-based EEG foundation model by keeping the backbone architecture fixed and modifying only the positional encoding component.
    
	\item We propose \textbf{Spherical Positional Encoding (SPE)}, which encodes electrode azimuth and inclination angles using multi-frequency sinusoidal mappings to incorporate spherical scalp geometry into transformer-based EEG models.

\end{itemize}


\mysection{materials and methods}

In this study, we conduct a controlled comparison of positional encoding strategies within the CBraMod \cite{cbramod2025} backbone. The original positional encoding module is replaced while all other architectural components, training protocols, and hyperparameters are kept identical, allowing us to isolate the specific contribution of positional encoding to representation learning and downstream generalization. To further assess robustness and adaptability, we evaluate downstream performance across multiple datasets with differing electrode montages and channel counts, enabling analysis under heterogeneous spatial configurations.

\mysubsection{Datasets} For pretraining the EEG foundation model, we used the Healthy Brain Network EEG dataset (HBN-EEG) \cite{shirazi2024hbn, alexander_open_2017, langer_eeg_2017}, which comprises recordings from more than 3,000 participants aged 5–21 years. EEG data were acquired using a 128-channel GSN-HydroCel montage. The dataset includes 11 releases from different participants. Recordings were collected during both passive paradigms (e.g., resting state, surround suppression, movie watching) and active tasks (e.g., contrast change detection, sequence learning, symbol search). We used the publicly available downsampled (100 Hz) and band-pass filtered (0.5–50 Hz) version of the dataset. The dataset already includes 60 Hz notch filtering in its released preprocessing pipeline. Releases 1–9 were used for pretraining and Release 10 for pretraining validation. Preprocessing steps included removing the reference channel, automatically detecting and interpolating bad channels \cite{noisy_ch, interpolate}, and applying average re-referencing. All preprocessing was performed using MNE-Python \cite{mne}. The pretrained models are evaluated on two downstream tasks: motor imagery classification and emotion recognition.

In motor imagery classification, we use the PhysioNet Motor Imagery (MI) dataset \cite{physionet2}. The dataset comprises EEG recordings from 109 subjects performing various motor and imagery tasks, acquired using 64 electrodes arranged according to the international 10–10 system at a sampling rate of 160 Hz. Only imagined movement trials are used, comprising four classes: left fist, right fist, both fists, and both feet.

For the emotion recognition task, we use the Fine-grained Affective Computing EEG Dataset (FACED) \cite{faced}, in which EEG recordings were collected from 123 participants watching 28 video clips covering nine emotional categories: anger, disgust, fear, sadness, amusement, inspiration, joy, tenderness, and neutral. Recordings were acquired using 32 electrodes following the international 10–20 system. We use the preprocessed version of the dataset, in which electrode ordering and naming are standardized across recording cohorts, the sampling rate is 250 Hz, and a band-pass filter of 0.05–47 Hz was applied. 


To standardize input lengths and control the number of transformer tokens, longer recordings were segmented into fixed-length 10-second epochs, while shorter trials, such as the original 4-second PhysioNet-MI trials, were used directly. The differences in frequency ranges across datasets arise from the publicly available preprocessed versions used in this study. No additional band-pass filtering was applied beyond the provided preprocessing pipelines. All datasets were resampled to 200 Hz for consistency with the backbone configuration. Subjects in the downstream datasets were split in a subject-independent manner, with 70\% assigned for training, 15\% for validation, and 15\% for testing. A random seed of 42 was used for dataset splitting.


Input data consist of multichannel EEG recordings $X \in \mathbb{R}^{C \times T}$, where $C$ denotes the number of electrodes  and $T$ represents the temporal samples. During pretraining and fine-tuning, each EEG epoch is further divided into non-overlapping 1-second windows, which are treated as input patches to the model. After patch extraction, each EEG sample is represented as $x \in \mathbb{R}^{C \times w \times t}$, where $t$ denotes the number of samples per patch and $w = \frac{T}{t}$ the number of patches per channel. Unlike the original CBraMod implementation, which normalizes signals by a fixed scalar of 100 after removing samples that exceed $100 \mu V$, we apply channel-wise max-absolute normalization to scale each channel to the range $[-1,1]$, as our pretraining dataset differs and comprises heterogeneous recordings with varying amplitude distributions across channels and subjects.


\[
\tilde{X}_{c,T} =
\frac{X_{c,T}}
{\max\left(\max_{T}|X_{c,T}|,\epsilon\right)}
\]

where $\epsilon$ is a small constant to prevent division by zero. This formulation preserves both zero-crossing structure and signal polarity. While normalization may affect positional encoding methods differently, the same preprocessing pipeline is used across all evaluated methods. For self-supervised pretraining, we adopt the random masking strategy of CBraMod, in which a predefined ratio of patch tokens is randomly selected and replaced with learnable mask token prior to being processed by the transformer backbone.

\mysubsection{CBraMod Architecture Overview} CBraMod is a self-supervised EEG foundation model designed to capture spatio-temporal dependencies via a criss-cross attention mechanism. The model first segments the EEG signal into non-overlapping patches. During pretraining, a random subset of patches is replaced with a learnable mask tokens. For each patch, temporal embeddings are extracted via convolutional layers and spectral embeddings are derived through a Fast Fourier Transform followed by a fully connected layer, and the two are fused via element-wise addition to obtain the final patch embedding. Positional encoding is then computed from the patch embeddings and added to form the input sequence to the transformer. In our experiments, the positional encoding module is systematically replaced to evaluate different strategies while keeping the rest of the architecture fixed.

The resulting sequence is processed by a criss-cross transformer backbone, in which attention is factorized into two operations: spatial attention, which models inter-channel relationships across electrodes, and temporal attention, which captures dependencies across time windows. This design enables the model to jointly learn spatial and temporal structure while maintaining computational efficiency.
The model is optimized via masked patch reconstruction, where the mean squared error (MSE) between the original and reconstructed masked patches is used as the training objective:

\[
L = \| \hat{X}_M - \tilde{X}_M \|_2^2
\]

where $\tilde{X}_M$ and $\hat{X}_M$ denote the original normalized patch and reconstructed masked patch, respectively.


\mysubsection{Positional Encoding Variants} \textbf{ACPE.} The Asymmetric Conditional Positional Encoding proposed in \cite{cbramod2025} dynamically generates positional encodings conditioned on patch embeddings via a depthwise two-dimensional convolution with an asymmetric kernel of size $(k_s, k_t)$, where $k_s > k_t$. The larger spatial kernel $k_s$ is designed to capture longer-range inter-channel positional relationships, while the smaller temporal kernel $k_t$ focuses on encoding local temporal positional information. The positional encoding is added to the original patch embedding before entering the transformer backbone.

\textbf{SPE.} While ACPE derives positional information implicitly from patch content, we propose Spherical Positional Encoding (SPE), which incorporates explicit anatomical priors by encoding the physical configuration of electrodes on the scalp. For each electrode, azimuth $\theta$ and inclination $\phi$ angles are obtained from its canonical spatial location. Each angular component is mapped through multi-frequency sine and cosine functions to obtain bounded and continuous representations that preserve periodicity:

$$pos_{c} =
\big[
\sin(\omega_i \theta), \cos(\omega_i \theta),
\sin(\omega_i \phi), \cos(\omega_i \phi)
\big]$$

where $\omega_i$ denotes the i-th frequency of the sinusoidal basis. By using multiple frequencies, the encoding captures angular relationships at different spatial scales. The resulting features are concatenated to form the channel positional embedding $pos_{c} = [P_{azimuth}, P_{inclination}]$, providing a continuous representation of global scalp topology that is invariant to head size scaling.

For temporal structure, standard sinusoidal positional encoding \cite{attention} is applied over patch indices to obtain patch positional encoding $pos_{w}$. The final SPE is obtained by summation:
$$ P_{cw} = pos_{w} + pos_{c} $$ 

This formulation jointly encodes spatial topology and temporal ordering without introducing additional learnable parameters.

As an extension, we additionally evaluate a projected variant (\textbf{SPE+Proj}), in which the channel and patch positional embeddings are each passed through separate linear projections before summation: 

\[
\mathrm{P}_{\text{cw}} = W_w \mathrm{pos}_w + W_{c} \mathrm{pos}_{c}
\]

where $W_w$ and $W_c$ are learnable projection matrices. This variant explores whether allowing the model to adaptively weight the fixed geometric and temporal encodings improves downstream performance.

\textbf{Learnable PE}. We also evaluate fully learnable positional encoding, in which both channel and patch positional embeddings are learned directly from data and summed to form the final positional encoding:
$$ P_{L} = e_{w}^{L} + e_{c}^{L} $$ 

where $e_{c}^{L} \in \mathbb{R}^{Cxd}$ and $e_{w}^{L} \in \mathbb{R}^{Wxd}$ are learnable embedding matrices for channel and patch indices, respectively. The learned channel embeddings are montage-specific and cannot be directly transferred across datasets with different electrode configurations. Since the pretraining and downstream datasets in our setup use different montages, both channel and patch embeddings are reinitialized and learned from scratch during fine-tuning.

\mysubsection{Implementation Details} We follow the same pretraining and fine-tuning procedure as CBraMod. All models are trained on 4× H100 80GB GPUs with 2× 24-core EPYC 9254 CPUs at 2.90 GHz and 1.5 TB RAM. Both pretraining and fine-tuning use 1-second non-overlapping patches as input.
The transformer backbone consists of 12 layers with 8 attention heads, an embedding dimension of $d=200$, and a feedforward dimension of 800. During pretraining, models are trained for 40 epochs with a batch size of 32 and a learning rate of $10^{-5}$ scheduled with cosine annealing and the AdamW optimizer. Additionally, 50\% of patch tokens are randomly selected and masked.

During fine-tuning, models are trained for 50 epochs with a batch size of 64 using the AdamW optimizer and label smoothing of 0.1. Weight decay during fine-tuning is set to 0.001. Model checkpoints are saved based on validation Cohen's Kappa, with higher Kappa indicating improved performance. All fine-tuning experiments are repeated with 5 different random seeds, and downstream performance is reported in terms of balanced accuracy, Cohen's Kappa, and weighted F1 score. For PhysioNet-MI fine-tuning, a three-layer classification head was used as it yielded better performance. For all other experiments, a single-layer head was used, as SPE exhibited training instability with a deeper classifier.

\begin{table*}[t]
\footnotesize
\setlength{\tabcolsep}{8pt}
\caption{Linear probe performance comparison of each positional encoding strategy on motor imagery (PhysioNet MI) and emotion recognition (FACED) tasks.}
\label{tab:linearprobe}
\centering
\begin{tabular}{lccc|ccc}
\hline
 & \multicolumn{3}{c|}{PhysioNet-MI (4-class)} & \multicolumn{3}{c}{FACED (9-class)} \\
\cline{2-7}
Model & Bal. Acc. & Cohen's Kappa & Weighted F1 & Bal. Acc. & Cohen's Kappa & Weighted F1 \\
\hline
NoPE & 0.4757 $\pm$ 0.0111 & 0.3010 $\pm$ 0.0148 & 0.4731 $\pm$ 0.0102
& 0.4098 $\pm$ 0.0082 & 0.3344 $\pm$ 0.0088 & 0.4056 $\pm$ 0.0078  \\
ACPE & 0.4993 $\pm$ 0.0065 & 0.3324 $\pm$ 0.0086 & 0.4940 $\pm$ 0.0069
     & \textbf{0.4212 $\pm$ 0.0053} & \textbf{0.3461 $\pm$ 0.0062} & \textbf{0.4180 $\pm$ 0.0065} \\
SPE & \textbf{0.5105 $\pm$ 0.0057} & \textbf{0.3477 $\pm$ 0.0077} & \textbf{0.5090 $\pm$ 0.0047}
    & 0.3878 $\pm$ 0.0061 & 0.3064 $\pm$ 0.0075 & 0.3766 $\pm$ 0.0084 \\
SPE+Proj. & 0.4835 $\pm$ 0.0111 & 0.3116 $\pm$ 0.0149 & 0.4841 $\pm$ 0.0107
& 0.3466 $\pm$ 0.0043 & 0.2625 $\pm$ 0.0032 & 0.3421 $\pm$ 0.0010 \\
\hline
\end{tabular}
\end{table*}

\begin{table*}[t]
\footnotesize
\setlength{\tabcolsep}{8pt}
\caption{Fine-tuning performance comparison of each positional encoding strategy on motor imagery (PhysioNet MI) and emotion recognition (FACED) tasks.}
\label{tab:finetune}
\centering
\begin{tabular}{lccc|ccc}
\hline
 & \multicolumn{3}{c|}{PhysioNet-MI (4-class)} & \multicolumn{3}{c}{FACED (9-class)} \\
\cline{2-7}
Model & Bal. Acc. & Cohen's Kappa & Weighted F1 & Bal. Acc. & Cohen's Kappa & Weighted F1 \\
\hline
NoPE & 0.5971 $\pm$ 0.0034 & 0.4628 $\pm$ 0.0045 & 0.5985 $\pm$ 0.0034 

& 0.4975 $\pm$ 0.0116 & 0.4306 $\pm$ 0.0124 & 0.4929 $\pm$ 0.0108 \\

ACPE & \textbf{0.6171 $\pm$ 0.0115} & \textbf{0.4892 $\pm$ 0.0153} & \textbf{0.6175 $\pm$ 0.0116}

     & 0.5013 $\pm$ 0.0139 & 0.4335 $\pm$ 0.0147 & 0.4930 $\pm$ 0.0116 \\
     
SPE & 0.6009 $\pm$ 0.0107 & 0.4676  $\pm$ 0.0143 & 0.6001 $\pm$  0.0109

    & 0.4820 $\pm$ 0.0070& 0.4136 $\pm$ 0.0083  & 0.4796 $\pm$ 0.0081\\
    
SPE+Proj. & 0.5983 $\pm$ 0.0149 & 0.4642 $\pm$ 0.0199 & 0.5992 $\pm$ 0.0143

& 0.4952 $\pm$ 0.0113 & 0.4289 $\pm$ 0.0120 & 0.4895 $\pm$ 0.0090 \\

Learnable PE & 0.6001 $\pm$ 0.0147 & 0.4666 $\pm$ 0.0197 & 0.6008 $\pm$ 0.0034 

& \textbf{0.5124 $\pm$ 0.0177} & \textbf{0.4467 $\pm$ 0.0184} & \textbf{0.5078 $\pm$ 0.0170} \\
\hline
\end{tabular}
\end{table*}


\mysection{results}

We evaluate the five positional encoding strategies under both linear probing and fine-tuning protocols on motor imagery classification and emotion recognition to assess how positional encoding design affects representation quality and downstream performance. Note that the results reported here are not directly comparable to those in the original CBraMod paper, as we use a different pretraining dataset and normalization strategy.

\begin{figure}[h]
	\centering
	\includegraphics[scale=0.5]{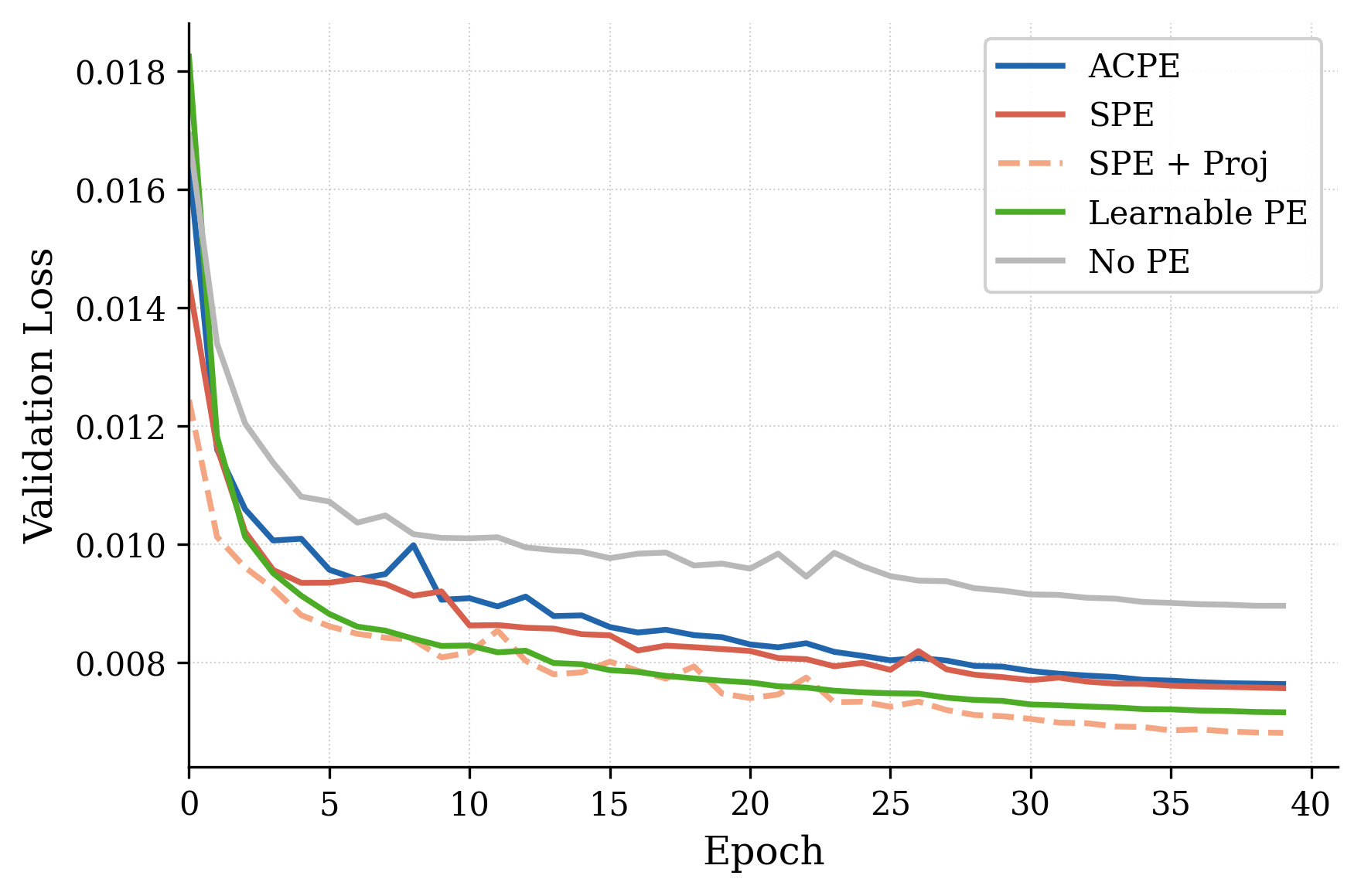}
	\caption{Pretraining masked reconstruction loss curves for CBraMod trained with five positional encoding strategies}
	\label{fig:loss}
\end{figure}

As shown in Figure \ref{fig:loss}, the absence of positional encoding results in the highest pretraining validation loss, demonstrating that positional information contributes to masked patch reconstruction. Among the remaining strategies, SPE converges faster than ACPE in early epochs but reaches a similar final loss. Learnable PE and SPE+Proj achieve the lowest and comparable validation losses.

Tab. \ref{tab:linearprobe} reports linear probe performance, where all backbone weights are frozen and only a single-layer classification head is trained to evaluate the quality of pretrained representations. Since learnable PE requires channel embeddings to be reinitialized due to montage mismatch between pretraining and downstream datasets, it is excluded from linear probe evaluation. SPE achieves the best performance on PhysioNet-MI, while ACPE yields the best results on FACED.

On the other hand, Tab. \ref{tab:finetune} shows fine-tuning performance, where all model weights are updated. For learnable PE, both channel and patch positional embeddings are reinitialized and learned from scratch during fine-tuning. ACPE achieves the best performance on PhysioNet-MI, while Learnable PE yields the best results on FACED.



\mysection{discussion}

No single positional encoding strategy achieves consistently superior performance across both tasks and evaluation protocols, suggesting that the optimal PE design is task- and dataset-dependent. Although the observed performance differences within each evaluation protocol are relatively small, all experiments differ only in the positional encoding module while keeping the remaining architecture fixed. Therefore, even modest improvements indicate measurable effects of positional information on learned EEG representations.

\textbf{NoPE} is consistently the weakest strategy in linear probing, confirming that positional information contributes meaningfully to representation quality during pretraining. However, NoPE remains competitive with several PE strategies after fine-tuning, indicating that the transformer backbone can partially recover positional structure through end-to-end optimization on the downstream task.

\textbf{SPE} produces competitive representations in the linear probe protocol for motor imagery, suggesting that explicit scalp topology encoding can be beneficial when spatial electrode layout is relevant to the task. One possible explanation is that motor imagery signals may exhibit more spatially distinguishable activation patterns, making the spherical geometry prior a useful inductive bias during pretraining. In contrast, SPE underperforms on FACED under both protocols, which may indicate that emotion-related EEG representations benefit less from a strict geometric prior. However, these observations remain hypothesis-driven and require further investigation. A practical advantage of SPE is that it requires no learned parameters and is precomputed from electrode angles, making it applicable to any montage without retraining.

\textbf{SPE+Proj} does not consistently improve over plain SPE despite introducing learnable projections. The learned projections likely corrupt the structured geometric prior rather than enhancing it.

\textbf{ACPE} either achieves the best performance or remains competitive across both tasks and protocols. By conditioning positional encodings on local patch content through asymmetric depthwise convolutions, ACPE captures both spatial and temporal neighborhood structure without relying on explicit electrode coordinates. This content-conditioned formulation appears more generalizable across diverse EEG tasks. However, since positional information is derived implicitly from patch representations rather than absolute electrode positions, ACPE is sensitive to channel ordering and performance may degrade if channels are permuted.

\textbf{Learnable PE} requires channel embeddings to be reinitialized during fine-tuning when the downstream montage differs from pretraining, which prevents direct transfer of spatial representations. Despite this, learnable PE achieves the best fine-tuning performance on FACED, suggesting that when the model is free to learn task-specific spatial structure from scratch on the target dataset, it adapts more effectively to the distributed spatial patterns characteristic of emotion recognition.



\mysection{conclusion}

In this paper, we benchmarked the effect of positional encoding design on transformer-based EEG foundation models by systematically replacing the positional encoding module in the CBraMod backbone and evaluating five strategies under linear probing and fine-tuning protocols on motor imagery and emotion recognition tasks. Our results confirm that positional information is beneficial for pretraining, as NoPE yields the highest masked reconstruction loss and consistently weak linear probe performance on downstream tasks.

No single positional encoding strategy achieves consistently superior performance across both tasks, with results varying based on task and dataset characteristics. SPE yields strong representations for motor imagery, where spatially distinguishable patterns may make the geometric prior beneficial, but is less effective for emotion recognition. Further experiments are required to better understand the interaction between task characteristics and positional encoding design. Additionally, SPE is precomputed, parameter-free, and montage-agnostic, making it a lightweight option when spatial structure is relevant. ACPE, by learning local neighborhood patterns from patch content, demonstrates more consistent performance across both tasks. When the pretraining and downstream montages differ, relearning positional embeddings from scratch on the target domain can outperform fixed strategies.

Evaluations are limited to a single backbone architecture and two downstream datasets; results may differ with alternative architectures or larger and more diverse evaluation benchmarks. Future work could explore PE strategies across more diverse EEG tasks and electrode montages, as well as larger pretraining datasets to better assess the generalizability of topology-aware encoding approaches.


\mysection{ACKNOWLEDGEMENTS}

This work was supported by the AI Co-Working Lab research initiative, funded by the Investitionsbank Sachsen-Anhalt through the Research and Innovation Program (EFRE – European Regional Development Fund).


\mysection{references}


${[1]}$ Fernandez J, Innocenti B, López B. EEG classification for neurological disorders using frequency band deciles. Scientific Reports. 2025;15:45142.

${[2]}$ Ávila CG, Bott FS, Tiemann L, Hohn VD, May ES, Nickel MM, et al. DISCOVER-EEG: an open, fully automated EEG pipeline for biomarker discovery in clinical neuroscience. Scientific Data. 2023;10:613.

${[3]}$ Guerrero-Mendez CD, Blanco-Diaz CF, Ruiz-Olaya AF, López-Delis A, Jaramillo-Isaza S, Milanezi Andrade R, et al. EEG motor imagery classification using deep learning approaches in naïve BCI users. Biomedical Physics \& Engineering Express. 2023;9(4).

${[4]}$ Torres EPP, Torres EA, Hernández-Álvarez M, Yoo SG. EEG-based BCI emotion recognition: A survey. Sensors. 2020;20(18):5083.

${[5]}$ Devlin J, Chang MW, Lee K, Toutanova K. BERT: Pre-training of deep bidirectional transformers for language understanding. In: Proc. NAACL. 2019, 4171–4186.

${[6]}$ Dosovitskiy A, Beyer L, Kolesnikov A, Weissenborn D, Zhai X, Unterthiner T, et al. An image is worth 16x16 words: Transformers for image recognition at scale. ICLR. 2021.

${[7]}$ Kostas D, Aroca-Ouellette S, Rudzicz F. BENDR: Using transformers and a contrastive self-supervised learning task to learn from massive amounts of EEG data. Frontiers in Human Neuroscience. 2021;15.

${[8]}$ Yang C, Westover M, Sun J. BIOT: Biosignal transformer for cross-data learning in the wild. In: NeurIPS. 2023, 78240–78260.

${[9]}$ Wang J, Zhao S, Luo Z, Zhou Y, Jiang H, Li S, et al. CBraMod: A criss-cross brain foundation model for EEG decoding. In: ICLR. 2025.

${[10]}$ Jiang W, Zhao L, Lu BL. Large brain model for learning generic representations with tremendous EEG data in BCI. In: ICLR. 2024.

${[11]}$ Cui W, Jeong W, Thölke P, Medani T, Jerbi K, Joshi AA, et al. Neuro-GPT: Towards a foundation model for EEG. In: IEEE ISBI. 2024, 1–5.

${[12]}$ Ouahidi YE, Lys J, Thölke P, Farrugia N, Pasdeloup B, Gripon V, et al. REVE: A foundation model for EEG – adapting to any setup with large-scale pretraining on 25,000 subjects. arXiv preprint arXiv:2510.21585. 2025.

${[13]}$ Klein T, Minakowski P, Sager S. Flexible patched brain transformer model for EEG decoding. Scientific Reports. 2025;15:10935.

${[14]}$ Shirazi SY, Franco A, Hoffmann MS, Esper NB, Truong D, Delorme A, et al. HBN-EEG: The FAIR implementation of the Healthy Brain Network (HBN) electroencephalography dataset. bioRxiv. 2024.

${[15]}$ Alexander LM, Escalera J, Ai L, Andreotti C, Febre K, Mangone A, et al. An open resource for transdiagnostic research in pediatric mental health and learning disorders. Scientific Data. 2017;4(1):170181.

${[16]}$ Langer N, Ho EJ, Alexander LM, Xu HY, Jozanovic RK, Henin S, et al. A resource for assessing information processing in the developing brain using EEG and eye tracking. Scientific Data. 2017;4(1):170040.

${[17]}$ Bigdely-Shamlo N, Mullen T, Kothe C, Su KM, Robbins KA. The PREP pipeline: Standardized preprocessing for large-scale EEG analysis. Frontiers in Neuroinformatics. 2015;9.

${[18]}$ Perrin F, Pernier J, Bertrand O, Echallier JF. Spherical splines for scalp potential and current density mapping. Electroencephalography and Clinical Neurophysiology. 1989;72(2):184–187.

${[19]}$ Gramfort A, Luessi M, Larson E, Engemann DA, Strohmeier D, Brodbeck C, et al. MEG and EEG data analysis with MNE-Python. Frontiers in Neuroscience. 2013;7.

${[20]}$ Schalk G. EEG motor movement/imagery dataset. PhysioNet. 2009.

${[21]}$ Chen J, Wang X, Huang C, Hu X, Shen X, Zhang D. A large finer-grained affective computing EEG dataset. Scientific Data. 2023;10(1):740.

${[22]}$ Vaswani A, Shazeer N, Parmar N, Uszkoreit J, Jones L, Gomez AN, et al. Attention is all you need. In: Proc. NeurIPS. 2017, 6000–6010.

\end{document}